\documentclass[conference]{IEEEtran}
\IEEEoverridecommandlockouts
\usepackage{cite}
\usepackage{amsmath,amssymb,amsfonts}
\usepackage{algorithmic}
\usepackage{graphicx}
\usepackage{textcomp}
\usepackage{xcolor}
\usepackage{arabtex}
\usepackage{utf8}
\usepackage{microtype}
\usepackage[hyphens]{url}
\usepackage{hyperref}
\hypersetup{breaklinks=true}
\def\BibTeX{{\rm B\kern-.05em{\sc i\kern-.025em b}\kern-.08em
    T\kern-.1667em\lower.7ex\hbox{E}\kern-.125emX}}
\begin{document}

\title{Sentiment Analysis of Persian-English Code-mixed Texts
}

\author{\IEEEauthorblockN{1\textsuperscript{st} Nazanin Sabri}
\IEEEauthorblockA{\textit{electrical and computer engineering} \\
\textit{University of Tehran}\\
Tehran, Iran \\
nazanin.sabri@ut.ac.ir}
\and
\IEEEauthorblockN{2\textsuperscript{nd} Ali Edalat}
\IEEEauthorblockA{\textit{electrical and computer engineering} \\
\textit{University of Tehran}\\
Tehran, Iran \\
ali.edalat@ut.ac.ir}
\and
\IEEEauthorblockN{3\textsuperscript{rd} Behnam Bahrak}
\IEEEauthorblockA{\textit{electrical and computer engineering} \\
\textit{University of Tehran}\\
Tehran, Iran \\
bahrak@ut.ac.ir}
}

\maketitle

\begin{abstract}
The rapid production of data on the internet and the need to understand how users are feeling from a business and research perspective has prompted the creation of numerous automatic monolingual sentiment detection systems. More recently however, due to the unstructured nature of data on social media, we are observing more instances of multilingual and code-mixed texts. This development in content type has created a new demand for code-mixed sentiment analysis systems. In this study we collect, label and thus create a dataset of Persian-English code-mixed tweets. We then proceed to introduce a model which uses BERT pretrained embeddings as well as translation models to automatically learn the polarity scores of these Tweets. Our model outperforms the baseline models that use Naïve Bayes and Random Forest methods. 
\end{abstract}

\begin{IEEEkeywords}
code-mixed language, sentiment analysis, Persian-English text
\end{IEEEkeywords}

\section{Introduction}
\label{sec:introduction}
Online social networking platforms place very few constraints on the type and structure of the textual content posted online. The unstructured nature of these content results in the creation of textual content which could be far from the original grammatical, syntactic, or semantic rules of the language they are written in \cite{arora2019character, gautam2014sentiment}. One deviation that has been observed quite often is the use of words from more than one language in the text \cite{barik2019normalization}. Commonly known as “code-mixed” text. These texts are written in language A but include one or more words of language B (either written in the official alphabets of language B or transliterated to language A). \\
In this study, we investigate code-mixed Persian-English data and perform sentiment analysis on these texts. The reason why sentiment analysis of these texts would differ from a text written purely in the Persian language is that the words containing the emotional energies of the text could be written in English which would make the Persian-only sentiment analysis models unable to produce the correct outputs. Grammatical differences could also render such monolingual models useless \cite{xu2015proceedings}. The difficulties of this task have been shown in other language pairs \cite{mandal2018preparing, banerjee2020limsi_upv}. Our reasoning behind choosing English as the second language is the prominence of the usage of this language overall but more importantly among Persian speakers.\\
Since Persian is a low resource language, in order to perform the aforementioned task, we first needed to create a dataset of texts and label those texts with the correct sentiment scores. Thus, we begin by using the Twitter API and searching for a list of “Finglish” \footnote{English words that are written using the Persian/Farsi alphabet} words through the use of the API. After the data has been collected, 2 annotators labeled the 3640 tweets completely. A third annotator was then added to the project to label the tweets on which the previous two annotators did not agree on. \\
After the dataset collection and creation was completed, an ensemble model was created and used to detect the sentiment scores of the texts.\\
The rest of this paper is structured as follows: Section \ref{sec:related_work} provides a brief overview of related work. Next, we look at our dataset in detail in Section \ref{sec:data}. Our models, as well as our text cleaning and preparation steps are described in Section \ref{sec:methods}. We then report our results in Section \ref{sec:results} and the study is concluded in Section \ref{sec:conclusion}. 

\section{Related Work}
\label{sec:related_work}
The prevalence of code-mixed textual data, due to the unstructured and uncontrolled nature of the web as well as social networks, have resulted in a focus on the topic throughout recent years, including multiple shared-tasks being defined on the subject \cite{patra2018sentiment, solorio2014overview, patwa2020sentimix, patwa2020semeval}. \\
One of the language pairs that has been the center of attention in code-mixed text analysis is Hindi-English \cite{srivastava2020iit, joshi2016towards, prabhu2016towards, sitaram2015sentiment, rudra2016understanding, jhanwar2018ensemble}. With the large population of multilingual individuals in India, such forms of texts have become quite common. Other language pairs (such as Bengali-English, Bambara-French, and Tamil-English), however, have been studied as well \cite{mandal2018preparing, konate2018sentiment, chakravarthi2020sentiment, chakravarthi2020corpus}.\\ 
Some studies attempt to solve the issue by hand engineering features which help in the task. In \cite{ghosh2017sentiment} various features (e.g. the number of code switches in the text) were introduced and employed in a multi-layer perceptron model. Number of sentiment and opinion lexicon, number of uppercase words, and POS tags are among some of the features which have been utilized. \\
Other studies try to find methods with less need for feature selection. For instance, it has been attempted to use cross-lingual word-embeddings \cite{singh2020sentiment} or subword embeddings \cite{prabhu2016towards} to accomplish the task. In the SemEval-2020 task on Spanglish and Hinglish \cite{patwa2020semeval} it is reported that BERT and ensemble methods are the most common and successful approaches. In \cite{singh2018automatic} an approach is introduced to help deal with different variations of the same word by substituting words with consideration to their context words. In \cite{aguilar2020lince} a benchmark for linguistic code-switching is presented, the aim of which is to enable evaluation of models. \\ 
To the best of our knowledge, the dataset annotated and presented as part of this study is the first dataset for the Code-mixed Persian-English sentiment analysis task. We also believe that there are no other studies on this specific subset of the topic available.

\section{Data}
\label{sec:data}
\setcode{utf8}

In this section, we describe the distributions and characteristics of our data. As described in Section \ref{sec:introduction}, our dataset consists of 3,640 tweets labeled with polarity values. Our dataset fields include the terms that were searched via the API that resulted in the tweets' retrieval, the text content of the tweets, the three labels assigned to each tweet, and the final label, which is calculated through majority voting.\\
We selected a list of 44 unique Finglish (English words transliterated to Persian) terms in order to collect this data. Some of the words in the list include: \<پرفکت> (perfect),
\<هپی> (happy), \<بورينگ> (boring), and \<سينگل> (single). \\
In our dataset, 69.2\% of the instances received unanimous labels by the first two annotators. A third annotator was then asked to label the rest of the data. In the resulting dataset, 15.7\% were labeled as positive and 59.7\% as negative (and the remaining 21.5\% were labeled as neutral). Table \ref{example-annotations} displays two examples of annotated data in our dataset. The majority of the data being negative can be explained by two facts: One is that due to the access restrictions of Twitter in Iran, only 9.24\% of Iranians use Twitter \cite{social_penetration:2020} and as a result the subset of users on the platform are mostly from a particular belief system \cite{khazraee2019mapping} which could result in the observed negative opinions. Another reason is that the data collection process was conducted in the last months of 2019 which included the beginning stages of the spread of the Coronavirus in Iran. Even though our keywords did not relate to the Coronavirus in any way the shift in spirit was observed in our data. We however believe that this issue reflects the current state of our society and thus the dataset can still be used for the task of code-mixed sentiment detection as the tweets do include the characteristics of code-mixed language and there are enough examples in the dataset to allow the model to learn attributes of polarities other than negative.\\ 
To preserve the privacy of the users, all user mentions in the texts have been replaced by @USERMENTION. However, since all tweets were public (at least at the time of collection) and were collected using the official Twitter API, Tweet IDs have been provided to allow use in future research should the need arise. This dataset is publicly available on GitHub. \footnote{\url{https://github.com/nazaninsbr/Persian-English-Code-mixed-Sentiment-Analysis}}

\begin{table}
\footnotesize
\centering
\caption{Example annotated tweets from our dataset}
\begin{tabular}{cc}
\hline
\textbf{Tweet} & \textbf{Label} \\ \hline
@USERMENTION \< زندگی خوشگل نيست، بورينگ و بی معنيه> & Negative\\
\<اوهوم هپی نيو آواتار و ازين حرفا بانو جان> & Positive\\
\hline
\end{tabular}
\label{example-annotations} 
\end{table}

\section{Method}
\label{sec:methods}

In this section we will go over our text processing and feature extraction, and data representation methods as well as our model. \\
In the text processing step we aim to create a vectorized representation of the textual input in order to be able to fit the data into our machine learning model. To do so we take the following steps:
\begin{enumerate}
\item Finding the non-Persian words in the sentence: By the definition of our task we know that the text we are processing includes non-Persian words, however, we do not know which words in the sentence they are. The reason why this knowledge could be useful is that knowing which words they are would allow us to use methods such as translation to convert them to their original language. In order to find these words, we use a dataset of Persian words collected from Wikipedia \footnote{\url{https://github.com/behnam/persian-words-frequency/blob/master/persian-wikipedia.txt}}. The huge collection of articles available on Wikipedia ensures that the most frequent words of the language would be in the list. We then check the existence of every word in the list and if it is not in the list we add it to our non-Persian word candidates. 
\item Translation: Next we translate the non-Persian words. First, we use an automatic tool, Yandex \footnote{\url{https://translate.yandex.com}}. This tool, however, faces difficulties when asked to translate Twitter specific slangs or expressions. Thus, we use common Twitter expression lists \footnote{\url{bit.ly/3h2qyNm}} and create a dictionary of our own. 
\item Embedding creation: To create an embedding for our textual data we used the pretrained multilingual BERT \cite{devlin2018bert} model that Google has provided on their GitHub page \footnote{\url{https://github.com/google-research/bert}}. Since the model is multilingual, it would allow for creation of embeddings for words from more than one language which fits nicely with our problem since code-mixed data includes instances of more than one language. Further since the model uses the idea of subword tokenization and embedding, it could allow for a better understanding of slangs or other non-common words that could be made up of better known subwords. 
\end{enumerate}

After our data passes these steps, it is fed into an ensemble model consisting of three Bidirectional Long Short-Term Memory (Bi-LSTM) networks.
\begin{itemize}
\item Our first network is a stacked Bi-LSTM network. The model is quite simple and aims to encode the general information available in the sentences. 
\item Our second model adds the attention mechanism to the Bi-LSTM network which enables the model to pay more attention to the most important words in the sentence. The attention mechanism also helps with the encoding of long-distance dependencies and information. 
\item Another method we use to make sure long-distance dependencies are accounted for and that information is not lost in our model is the use of pooling layers in our final model. 
\end{itemize}
The final model takes the outputs of all three models and uses a weighted average to produce the output. To find the best weight assigned to the output of each model, we use the optimization algorithms offered by SciPy \cite{2020SciPy-NMeth}. 

\section{Results}
\label{sec:results}
We used 10-fold cross-validation and averaged the metrics in order to present more reliable values. Additionally, to be able to compare our results, naive Bayes and random forest models were also used on the data. The results have been presented in Table~\ref{results-table}. 

\begin{table}
\footnotesize
\centering
\caption{10-fold cross-validation model performance results: each reported metric is calculated by averaging the values across all 10 folds}
\begin{tabular}{ccccc}
\textbf{Model Name} & \textbf{Accuracy} & \textbf{Precision} & \textbf{Recall} & \textbf{F1}\\ \hline
Naive Bayes & 60.88 & 69.00 & 60.88 & 47.17\\
Random Forest & 62.12 & 60.37 & 62.12 & 57.67\\
Our Model & 66.17 & 64.16 & 65.99 & 63.66\\
\hline
\end{tabular}
\label{results-table} 
\end{table}

We can see that our ensemble model outperforms the baseline models with regards to all metrics. Through our experiments, we find that the attention and pooling mechanisms both help with the performance of the models. We further find that the sum of all three models, offers better performance than each individual model, as each model appears to make up for the shortcomings of the other models in the ensemble. 

\section{Conclusion}
\label{sec:conclusion}
In this study we presented a Persian-English code-mixed dataset. The dataset consisted of 3640 tweets collected through the use of the Twitter API. Each tweet was consequently labeled with its corresponding polarity score. We then used neural classification models to learn the polarity scores of these data. Our models employed Yandex and dictionary-based translation techniques to translate the code-mixed words in our texts. We further used pretrained BERT embeddings to represent our data. Our models reached an accuracy of 66.17\% and F1 of 63.66 on the data. \\
Future work could focus on other methods of dealing with the code-mixed words or ways in which we could find word-based polarity scores for our code-mixed words using sentence level scores. 

\bibliographystyle{unsrt}  
\bibliography{aacl-ijcnlp2020}

\end{document}